\documentclass[10pt,twocolumn,letterpaper]{article}

\usepackage{iccv}
\usepackage{times}
\usepackage{epsfig}
\usepackage{graphicx}
\usepackage{amsmath}
\usepackage{amssymb}
\usepackage{threeparttable}
\usepackage{multirow}
\usepackage{multicol}
\usepackage{booktabs}
\usepackage{adjustbox}
\usepackage{setspace}
\usepackage{authblk}

\usepackage[breaklinks=true,bookmarks=false]{hyperref}

\iccvfinalcopy 

\def\iccvPaperID{****} 

\begin{document}

\title{Rethinking on Multi-Stage Networks for Human Pose Estimation}

\author{
	Wenbo Li$^{1,2}$\thanks{The first two authors contribute equally to this work. This work is done when Wenbo Li, Binyi Yin, Qixiang Peng, Yuming Du and Tianzi Xiao are interns at Megvii Research.} \quad Zhicheng Wang$^{1*}$ \quad Binyi Yin$^{1}$ \quad Qixiang Peng$^{1}$ \quad Yuming Du$^{1,3}$ \quad Tianzi Xiao$^{1,4}$ \quad Gang Yu$^{1}$ \quad Hongtao Lu$^{2}$ \quad Yichen Wei$^{1}$ \quad Jian Sun$^{1}$ \vspace{0.10cm} \\
	$^1$Megvii Inc. (Face++)   
	$^2$Shanghai Jiao Tong University
	$^3$Beihang University \\
	$^4$Beijing University of Posts and Telecommunications \\
	$^{1}$\textit{\{liwenbo,wangzhicheng,yinbinyi,pengqixiang,duyuming,xiaotianzi,yugang,weiyichen,sunjian\}@megvii.com}
	$^{2}$ \textit{htlu@sjtu.edu.cn}
}

\maketitle

\begin{abstract}
	Existing pose estimation approaches fall into two categories: \emph{single-stage} and \emph{multi-stage} methods. While multi-stage methods are seemingly more suited for the task, their performance in current practice is not as good as single-stage methods.
	
	This work studies this issue. We argue that the current multi-stage methods' unsatisfactory performance comes from the insufficiency in various design choices. We propose several improvements, including the single-stage module design, cross stage feature aggregation, and coarse-to-fine supervision. The resulting method establishes the new state-of-the-art on both MS COCO and MPII Human Pose dataset, justifying the effectiveness of a multi-stage architecture. The source code is publicly available for further research.\footnote{https://github.com/megvii-detection/MSPN}
\end{abstract}

\section{Introduction}
Human pose estimation problem has seen rapid progress in recent years using deep convolutional neural networks. Currently, the best performing methods~\cite{papandreou2017towards,he2017mask,chen2018cascaded,xiao2018simple} are pretty simple, typically based on a \emph{single-stage} backbone network, which is transferred from image classification task. For example, the COCO keypoint challenge 2017 winner~\cite{chen2018cascaded} is based on Res-Inception~\cite{szegedy2017inception}. The recent simple baseline approach~\cite{xiao2018simple} uses ResNet~\cite{he2016deep}. As pose estimation requires a high spatial resolution, up sampling~\cite{chen2018cascaded} or deconvolution~\cite{xiao2018simple} is appended after the backbone networks to increase the spatial resolution of deep features.

\begin{figure}
	\begin{center}
		\includegraphics[width=0.9\linewidth]{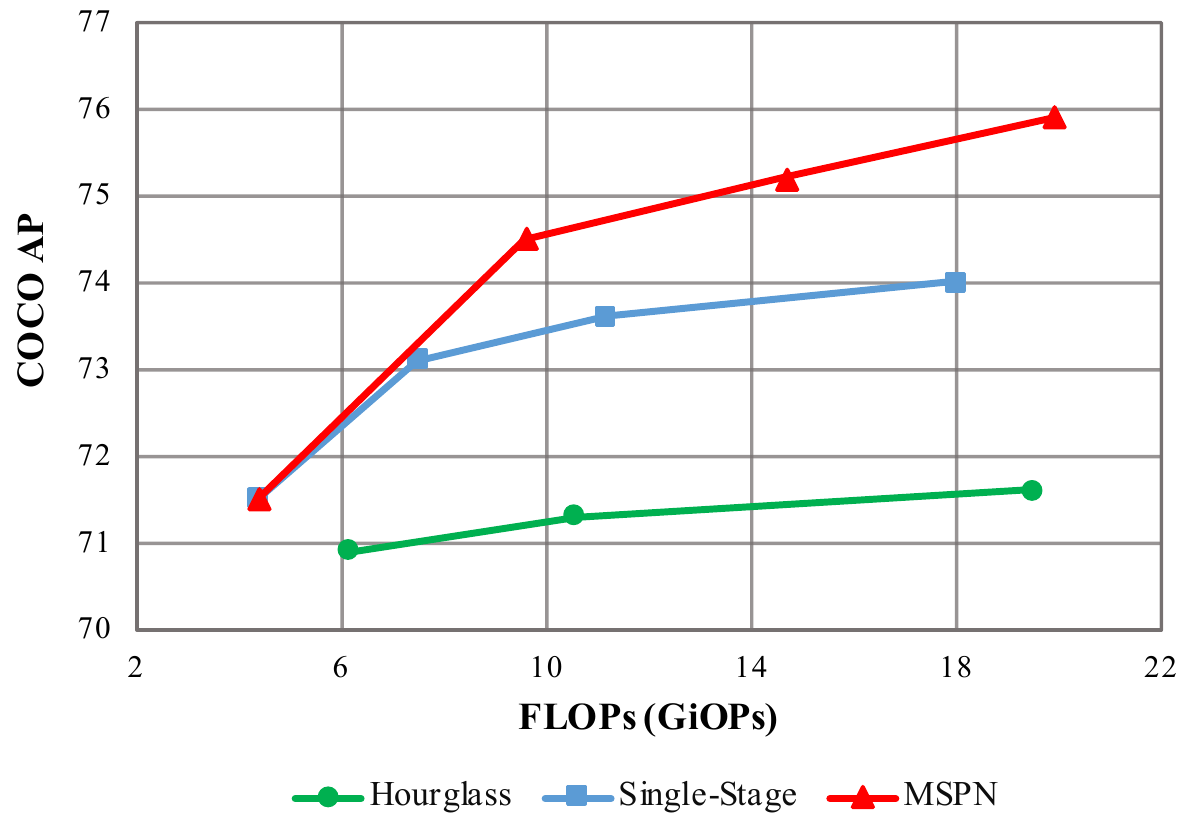}
	\end{center}
	\caption{Pose estimation performance on COCO minival dataset of Hourglass~\cite{newell2016stacked}, a single-stage model using ResNet~\cite{he2016deep}, and our proposed MSPN under different model capacity (measured in FLOPs).}
	\label{fig:FLOPs}
\end{figure}

Another category of pose estimation methods adopts an \emph{multi-stage} architecture. Each stage is a simple light-weight network and contains its own down sampling and up sampling paths. The feature (and heat) maps between the stages remain a high resolution. All the stages are usually supervised simultaneously to facilitate a coarse-to-fine, end-to-end training. Representative works include convolutional pose machine~\cite{wei2016convolutional} and Hourglass network~\cite{newell2016stacked}.

At an apparent look, the multi-stage architecture is more suited for the pose estimation task because it naturally enables high spatial resolution and is more flexible. Indeed, multi-stage methods are dominant on MPII~\cite{andriluka20142d} dataset (mostly the variants of Hourglass~\cite{newell2016stacked}). However, they are not as good as single-stage methods on the more challenging COCO dataset. Based on the previous works, it is unclear \emph{whether a multi-stage architecture is better or not}.

This work aims to study this issue. We point out that the current unsatisfactory performance in multi-stage methods is mostly due to the insufficiency in various design choices. We show that the potential advantage of a multi-stage architecture can be better exploited with certain improvements on those design choices, and SOTA results can be achieved. 

Specifically, we propose a multi-stage pose estimation network (MSPN) with three improvements. \emph{First}, we notice that the single-stage module in the current multi-stage methods is not good. For example, a Hourglass~\cite{newell2016stacked} module uses equal width channels in all blocks for both down and up sampling. Such a design is clearly inconsistent with the current good practice in network architecture design such as ResNet~\cite{he2016deep}. We found that simply adopting the existing good network structure (GlobalNet of CPN~\cite{chen2018cascaded} in this work) as the single-stage module is sufficiently good. \emph{Second}, due to the repeated down and up sampling steps, information is more likely to lose and optimization becomes more difficult. We propose to aggregate features across different stages to strengthen the information flow and mitigate the difficulty in training. \emph{Last}, observing that the pose localization accuracy is gradually refined during multi-stage, we adopt a coarse-to-fine supervision strategy in accordance. Note that this is different from the commonly used multi-scale supervision in previous works~\cite{newell2016stacked, newell2017associative, yang2017learning, ke2018multi}.

With above improvements playing in synergy, the resulting multi-stage architecture significantly outperforms all previous works. This is exemplified in Figure~\ref{fig:FLOPs}. For the single-stage method, its performance becomes saturated while increasing the network capacity. As shown in Table~\ref{tab:single}, Res-152 obtains 73.6 AP on COCO minival dataset and Res-254 has 74.0, only 0.4 improvement. For the representative multi-stage method Hourglass~\cite{newell2016stacked}, only a small performance gain is obtained after using more than two stages. As illustrated in Table~\ref{tab:multi}, it has 71.3 AP at 4 stages and 71.6 at 8 stages, only 0.3 improvement. With similar FLOPs, MSPN has 74.5 AP at 2 stages and 75.9 at 4 stages, that is 1.4 point improvement. Therefore, MSPN has a clearly better accuracy-FLOPs tradeoff.

New state-of-the-art performance is achieved. On COCO keypoint benchmark, the proposed single model achieves 76.1 average precision (AP) on test-dev. It significantly outperforms state-of-the-art algorithms. Finally, we obtain 78.1 AP on test-dev and 76.4 on test-challenge dataset, which is 4.3 AP improvement on test-challenge benchmark compared with MS COCO 2017 Challenge winner. Meanwhile, the proposed method obtains 92.6 PCKh@0.5 on MPII test dataset, which is also the best.

\section{Related Work}
Pose estimation has undergone a long way as a primary research topic of computer vision. In the early days, hand-crafted features are widely used in classical methods~\cite{andriluka2009pictorial,sapp2013modec,gkioxari2013articulated,sapp2010adaptive,dantone2013human,yang2011articulated,johnson2011learning,pishchulin2013poselet}. Recently, many approaches~\cite{bulat2016human,gkioxari2016chained,pishchulin2016deepcut,insafutdinov2016deepercut,carreira2016human,belagiannis2017recurrent} take advantage of deep convolutional neural network (DCNN)~\cite{krizhevsky2012imagenet} to enhance the performance of pose estimation by a large step. In terms of network architecture, current human pose estimation methods could be divided as single-stage~\cite{papandreou2017towards,he2017mask,chen2018cascaded,xiao2018simple} and multi-stage~\cite{wei2016convolutional,cao2016realtime,newell2017associative,newell2016stacked,yang2017learning,ke2018multi} two categories.

{\bfseries Single-Stage Approach} 
Single-stage methods~\cite{papandreou2017towards,he2017mask,chen2018cascaded,xiao2018simple} are based on backbone networks that are well tuned on image classification tasks, such as VGG~\cite{simonyan2014very} or ResNet~\cite{he2016deep}. Papandreou~\etal ~\cite{papandreou2017towards} designs a network to generate heat maps as well as their relative offsets to get the final predictions of the key points. He \etal ~\cite{he2017mask} proposes Mask R-CNN to first generate person box proposals and then apply single-person pose estimation. Chen~\etal ~\cite{chen2018cascaded} which is the winner of COCO 2017 keypoint challenge leverages a Cascade Pyramid Network (CPN) to refine the process of pose estimation. The proposed online hard key points mining (OHKM) loss is used to deal with hard key points. Xiao~\etal~\cite{xiao2018simple} provides a baseline method that is simple and effective in the pose estimation task. 
In spite of their good performance, these methods have encountered a common bottleneck. Simply increasing the model capacity does not give rise to much improvement in performance. This is illustrated in both Figure~\ref{fig:FLOPs} and Table~\ref{tab:single}.

{\bfseries Multi-Stage Approach} Multi-Stage methods\cite{wei2016convolutional,cao2016realtime,newell2017associative,newell2016stacked,yang2017learning,ke2018multi} aim to produce increasingly refined estimation. 
They can be bottom-up or top-down. In contrary, single-stage methods are all top-down.

Bottom-up methods firstly predict individual joints in the image and then associate these joints into human instances.  Cao~\etal~\cite{cao2016realtime} employs a VGG-19 ~\cite{simonyan2014very} network as a feature encoder, then the output features go through a multi-stage network resulting in heat maps and associations of the key points. Newell~\etal~\cite{newell2017associative} proposes a network to simultaneously output key points and group assignments.

\begin{figure*}[t]
	\centering
	\includegraphics[width=1.0\linewidth]{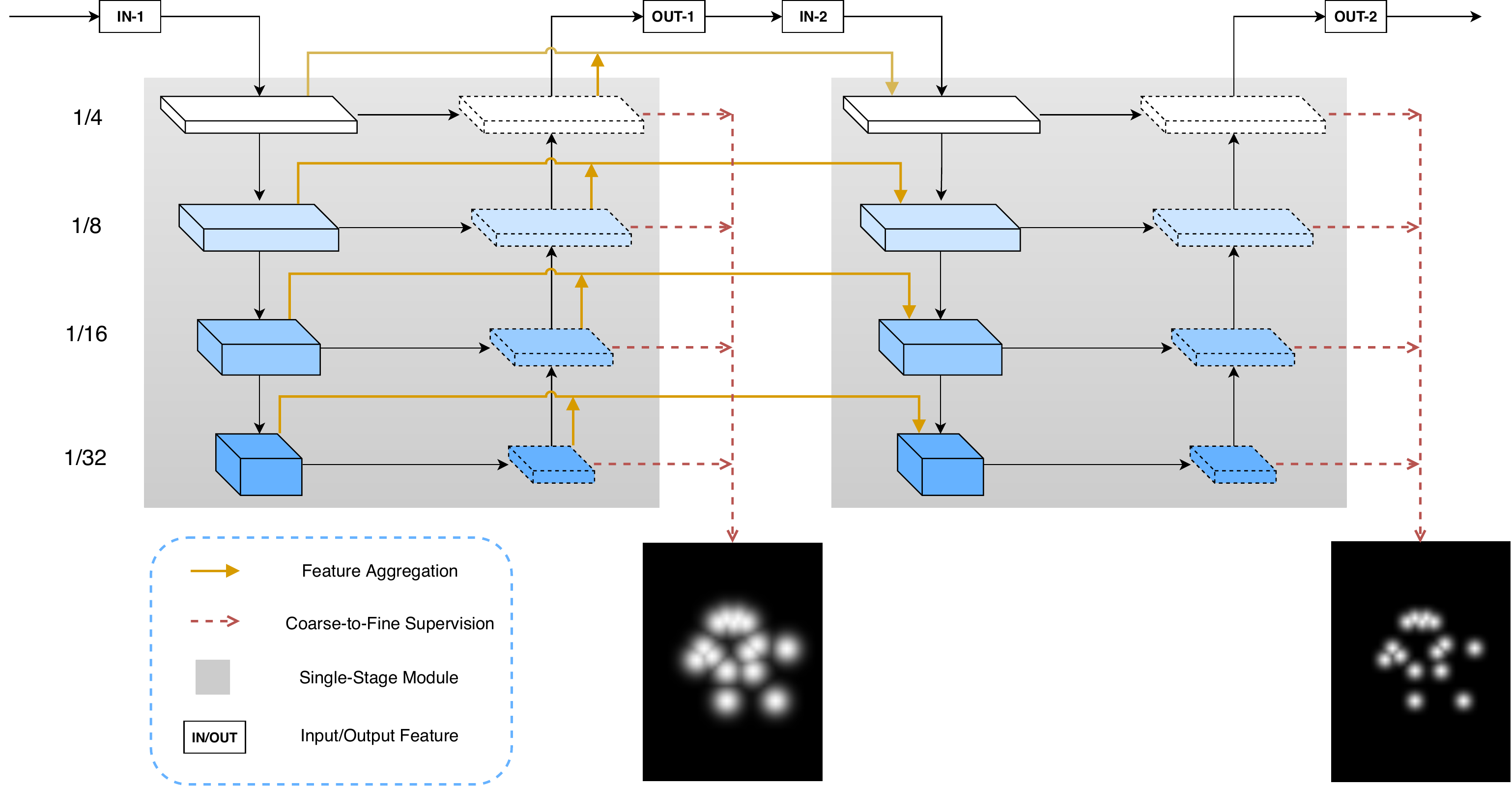}
	\caption{Overview of Multi-Stage Pose Network~(MSPN). It is composed of two single-stage modules. A cross stage aggregation strategy~(zoomed in Figure~\ref{fig:aggr}) is adopted between  adjacent stages~(Section~\ref{section:CSFA}). A coarse-to-fine supervision strategy further improves localization accuracy~(Section~\ref{setion:C2F}).}
	\label{fig:mspn}
\end{figure*}

Top-down approaches first locate the persons using detectors~\cite{ren2015faster,lin2017feature,li2018detnet}. And a single person pose estimator is then used to predict the key points' locations. Wei~\etal~\cite{wei2016convolutional} employs deep convolutional neural networks as feature encoder to estimate human pose. This work designs a sequential architecture composed of convolutional networks to implicitly model long-range dependencies between joints. Hourglass~\cite{newell2016stacked} is proposed to apply intermediate supervision to repeated down sampling, up sampling processing for pose estimation task. ~\cite{yang2017learning} adopts Hourglass and further design a Pyramid Residual Module (PRMs) to enhance the invariance in different scales. Many recent works~\cite{ke2018multi, Chen_2017_ICCV, Xiao-Multi, Tang_2018_ECCV, zhang2019human} are based on Hourglass and propose various improvements. 
While these multi-stage methods work well on MPII~\cite{andriluka20142d}, they are not competitive on the more challenging tasks on COCO~\cite{lin2014microsoft}. For example, the winners of COCO keypoint challenge on 2016~\cite{papandreou2017towards}, 2017~\cite{chen2018cascaded} are all single-stage based, as well as the recent simple baseline work~\cite{xiao2018simple}. In this work, we propose several modifications on existing multi-stage architecture and show that the multi-stage architecture is better.

\section{Multi-Stage Pose Network}
We adopt the top-down approach in two steps. In the first step, an off-the-shelf human detector is adopted. Quantitive comparison of different detectors in the experiments shows that the quality of the detector is inconsequential, as long as it is sufficiently good.

In the second step, the proposed Multi-Stage Pose Network~(MSPN) is applied to each detected human bounding box to produce the pose result. The network is exemplified in two stages as in Figure~\ref{fig:mspn}.

The Multi-Stage Pose Network proposes three improvements. First, we analyze the deficiency of the previous single-stage module and show why the state-of-the-art single-stage pose network can be readily exploited. Second, to reduce information loss, a feature aggregation strategy is proposed to propagate information from early stages to the later ones. Last, we introduce the usage of coarse-to-fine supervision. It adopts finer supervision in localization accuracy in later stages. 

The following sections elaborate on each improvement.

\subsection{Analysis of a Single-Stage Module}
\label{part:single_stage_module}

Most recent multi-stage methods~\cite{yang2017learning,ke2018multi, Chen_2017_ICCV, Xiao-Multi, Tang_2018_ECCV} are variants of Hourglass~\cite{newell2016stacked}. In each module of Hourglass, the number of convolutional filters (or feature maps) remains constant during repeated down and up sampling steps. This equal-channel-width design results in a relatively poor performance seen from Figure~\ref{fig:FLOPs} since a lot of information will be lost after every down sampling. 

By contrast, modern network architectures~\cite{simonyan2014very,he2016deep,xie2017aggregated,hu2017squeeze,chollet2017xception,howard2017mobilenets} are different. The number of feature maps is increased when there is a down sampling. Likewise, we note that there are some variants of Hourglass using different width channels. \cite{newell2017associative} uses the same number of channels $\left[256, 386, 512, 768 \right]$ in both down and up sampling paths. However, this variant has only 71.7 AP with 15.4G FLOPs at 2 stages as shown in Table~\ref{tab:multi}. Compared with this setting, our proposed MSPN remains a small number of feature maps $\left[256, 256, 256, 256 \right]$ during the up sampling and allocates more computation complexity to the down sampling. In it, the number of feature maps is doubled after every spatial down sampling. It is reasonable since we aim to extract more representative features in the down sampling process and the lost information can hardly be recovered in the up sampling procedure. Therefore, increasing the capacity of down sampling unit is usually more effective. Finally, 2-stage MSPN obtains 74.5 AP with 9.6G FLOPs.


In this work, we adopt the ResNet-based GlobalNet of CPN~\cite{chen2018cascaded} as the single-stage module. As shown in Figure~\ref{fig:mspn}, it is a U-shape architecture in which features extracted from multiple scales are utilized for predictions. Note that the single stage module structure itself is not novel, but applying it in a multi-stage setting is new and shown effective in this work for the first time. In the Section~\ref{part:multi-stage}, we also demonstrate that this module structure is general. The down sampling unit can effectively use other backbones as well.


\subsection{Cross Stage Feature Aggregation}\label{section:CSFA}
\label{part:csfa}
A multi-stage network is vulnerable by the information losing during repeated up and down sampling. To mitigate this issue, a cross stage feature aggregation strategy is used to propagate multi-scale features from early stages to the current stage in an efficient way.

\begin{figure}
	\centering
	\includegraphics[width=1.0\linewidth]{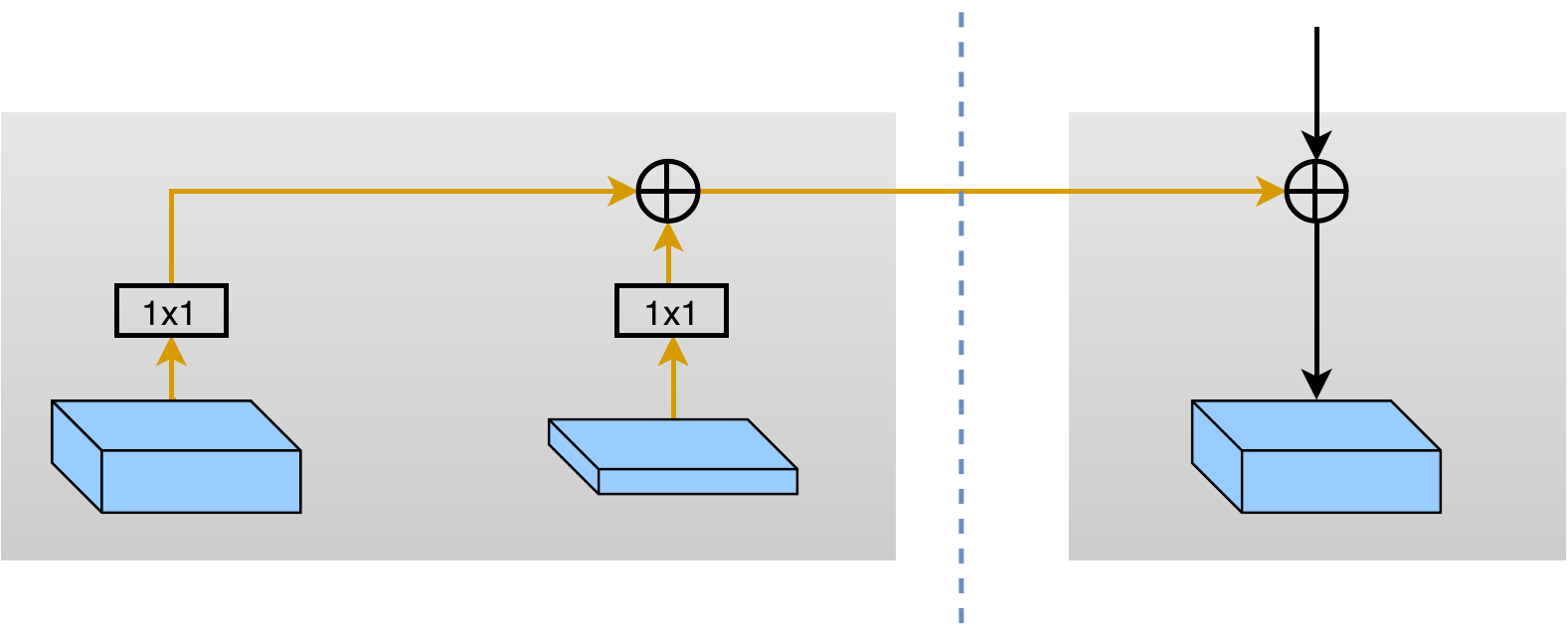}
	\caption{Cross Stage Feature Aggregation on a specific scale. Two $1\times1$ convolutional operations are applied to the features of previous stage before aggregation. See Figure~\ref{fig:mspn} for the overall network structure.}
	\label{fig:aggr}
\end{figure}

As is shown in Figure~\ref{fig:mspn}, for each scale, two separate information flows are introduced from down sampling and up sampling units in the previous stage to the down sampling procedure of the current stage. It is noted that a $1 \times 1$ convolution is added on each flow as shown in Figure~\ref{fig:aggr}. Together with down-sampled features of current stage, three components are added to produce fused results. With this design, the current stage can take full advantage of prior information to extract more discriminative representations. In addition, the feature aggregation could be regarded as an extended residual design, which is helpful dealing for with the gradient vanishing problem. 

\subsection{Coarse-to-fine Supervision}\label{setion:C2F}

\begin{figure}
	\centering
	\includegraphics[width=0.6\linewidth]{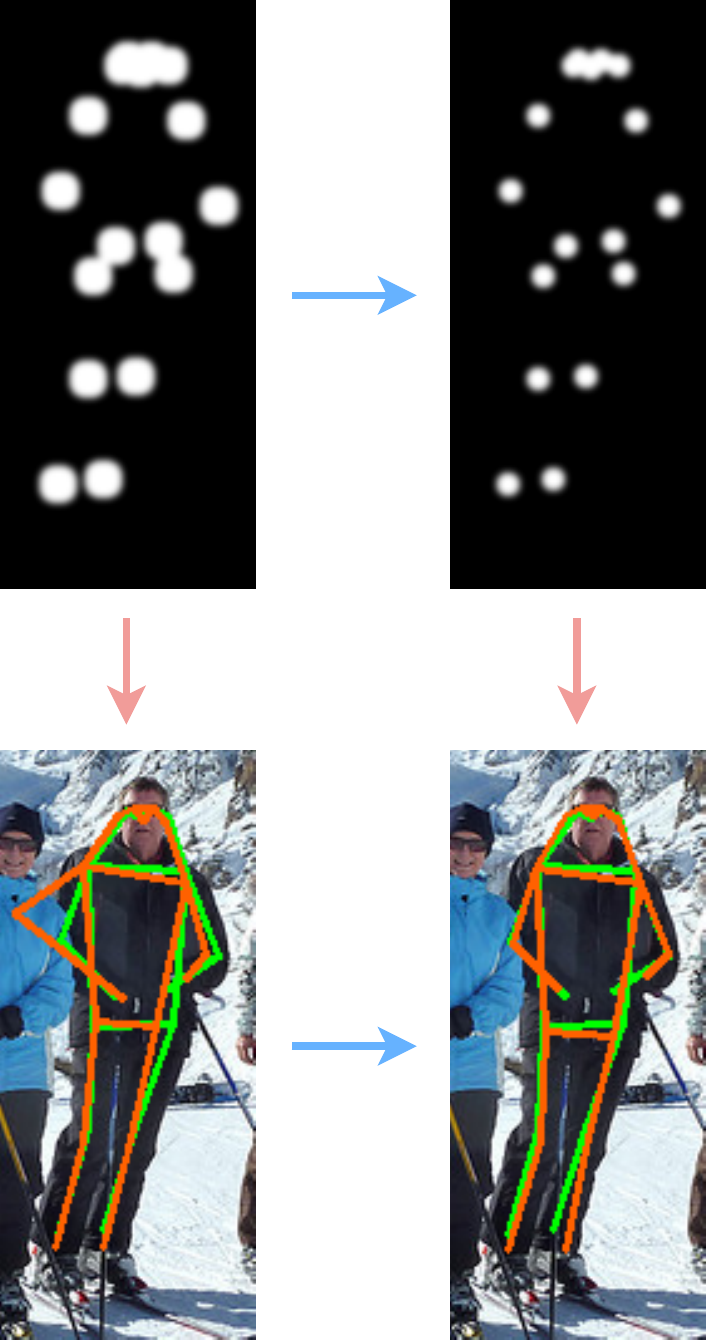}
	\caption{Illustration of coarse-to-fine supervision. The first row shows ground-truth heat maps in different stages and the second row represents corresponding predictions and ground truth annotations. The orange line is the prediction result and the green line indicates ground truth. }
	\label{fig:ctf}
\end{figure}

In the pose estimation task, context is crucial for locating the challenging poses since it provides information for invisible joints. Besides, we notice that small localization errors would seriously affect the performance of pose estimation. Accordingly, we design a coarse-to-fine supervision, as illustrated in Figure~\ref{fig:mspn}. 
Specifically, the ground truth heat map for each joint is realized as a Gaussian in most previous works. In this work, we further propose to use different kernel sizes of the Gaussian in different stages. That is, an early stage uses a large kernel and a latter stage uses a small kernel. This strategy is based on the observation that the estimated heat maps from multi-stages are also in a similar coarse-to-fine manner. Figure~\ref{fig:ctf} shows an illustrative example. It demonstrates that the proposed supervision is able to refine localization accuracy gradually.

Besides, we are inspired that intermediate supervision could play an essential role in improving the performance of deep neural network from~\cite{szegedy2015going}. Therefore, we introduce a multi-scale supervision to perform intermediate supervisions with four different scales in each stage, which could obtain substantial contextual information in various levels to help locate challenging poses. As shown in Figure~\ref{fig:mspn}, an online hard key points mining~(OHKM)~\cite{chen2018cascaded} is applied to the largest scale supervision in each stage. L2 loss is used for heat maps on all the scales.

\section{Experiments}

\subsection{Dataset and Evaluation Protocol}
MS COCO~\cite{lin2014microsoft} is adopted to evaluate the performance of our framework. It consists of three splits: train, validation and test. Similar to~\cite{chen2018cascaded}, we aggregate the data of train and validation parts together, and further divide it into trainval dataset (nearly 57K images and 150K person instances) and minival dataset (5k images). They are separately utilized for training and evaluating. OKS-based mAP (AP for short) is used as our evaluation metric~\cite{lin2014microsoft}.

MPII human Pose dataset~\cite{andriluka20142d} provides around 25k images from a variety of real-world activities. There are over 40k person instances with annotated body joints, among which 12k instances are used for testing and others for training. PCKh@0.5 is used to evaluate the performance of single-person pose estimation.

\subsection{Implementation Details}

{\bfseries Human Detector.}
We adopt a state-of-the-art object detector MegDet~\cite{peng2018megdet} to generate human proposals. The MegDet is trained with full categories of MS COCO dataset. Only human boxes out of the best 100 ones of all categories are selected as the input of single-person pose estimator. All the boxes are expanded to have a fixed aspect ratio of $4 \colon 3$ for COCO.

{\bfseries Training.}
The network is trained on 8 Nvidia GTX 1080Ti GPUs with mini-batch size 32 per GPU. There are 90k iterations. Adam optimizer is adopted and the linear learning rate gradually decreases from 5e-4 to 0. The weight decay is 1e-5. 

Each image will randomly go through a series of data augmentation operations including cropping, flipping, rotation, and scaling. As for cropping, instances with more than eight joints will be cropped to upper or lower bodies with equal possibility. The rotation range is $-45^{\circ} \sim 45^{\circ}$, and scaling range is $0.7 \sim 1.35$. The image size is set $256 \times 192$ in Section~\ref{part:ablation} and $384 \times 288$ in Section~\ref{part:sota} for MS COCO dataset, and $256 \times 256$ for MPII. 

{\bfseries Testing.}
A post-Gaussian filter is applied to the estimated heat maps. Following the same strategy as ~\cite{newell2016stacked}, we average the predicted heat maps of original image with results of corresponding flipped image. Then, a quarter offset in the direction from the highest response to the second highest response is implemented to obtain the final locations of key points. The pose score is the multiplication of box score and average score of key points, same as in ~\cite{chen2018cascaded}.

\subsection{Ablation Study}
\label{part:ablation}
In this section, we provide an in-depth analysis of each individual design in our framework.

In order to show the effectiveness of our method in a clear way, we also perform corresponding experiments on Hourglass~\cite{newell2016stacked}. All results are reported on COCO minival dataset. The input image size is $256 \times 192$.

\subsubsection{Multi-Stage Architecture}
\label{part:multi-stage}


First, we evaluate how the capacity of backbone affects the performance of pose estimation. In terms of the single-stage network in Table~\ref{tab:single}, we observe that its performance gets quickly saturated with the growth of backbone capacity. It is obvious that Res-101 outperforms Res-50 by 1.6 AP and costs a more 3.1G FLOPs, but there is only 0.5 gain from Res-101 to Res-152 at the cost of additional 3.7G FLOPs. For further exploration, we train a Res-254 network by adding more residual blocks on Res-152. Although the FLOPs of the network increases from 11.2G to 18.0G, there is an only 0.4 AP improvement. Therefore, it is not effective to adopt Res-152 or larger backbones for a single-stage network. 

\begin{table}[h]
	\begin{spacing}{1.1}
		\begin{center}
			\begin{tabular}{c|c c c c}
				\hline
				Method & Res-50 & Res-101 & Res-152 & Res-254 \\
				\hline
				AP & 71.5 & 73.1 & 73.6 & 74.0 \\
				FLOPs(G) & 4.4 & 7.5 & 11.2 & 18.0 \\
				\hline
			\end{tabular}
		\end{center}
		\caption{Results of single-stage networks with different backbones on COCO minival dataset.}
		\label{tab:single}
	\end{spacing}
\end{table}

\begin{table}[h]
	\begin{center}
		\renewcommand\tabcolsep{4.0pt}
		\begin{threeparttable}
			\begin{tabular}{c|c c||c|c c}
				\hline
				\multirow{2}*{\bf{Stages}} & \multicolumn{2}{c||}{\bf Hourglass} & \multirow{2}*{\bf{Stages}} & \multicolumn{2}{c}{\bf MSPN} \\ 
				~ & FLOPs(G) & AP & ~ & FLOPs(G) & AP \\
				\hline
				1 & 3.9 & 65.4 & 1 & 4.4 & 71.5 \\
				2 & 6.2 & 70.9 & 2 & 9.6 & 74.5 \\
				4 & 10.6 & 71.3 & 3 & 14.7 & 75.2 \\
				8 & 19.5 & 71.6 & 4 & 19.9 & 75.9 \\
				\hline
				2$^\dagger$ & 15.4$^\dagger$ & 71.7$^\dagger$ & - & - & - \\
				\hline
			\end{tabular}
		\end{threeparttable}
	\end{center}
	\caption{Results of Hourglass and MSPN with different number of stages on COCO minival dataset. "$\dagger$" denotes the result of a variant Hourglass~\cite{newell2017associative} as illustrated in Section ~\ref{part:single_stage_module}. MSPN adopts Res-50 in each single-stage module.}
	\label{tab:multi}
\end{table}

Then, we demonstrate the effectiveness of multi-stage architecture based on the proposed single-stage module. From Table~\ref{tab:multi}, we can see that the performance of single-stage Hourglass~\cite{newell2016stacked} is poor. Adding one more stage introduces a large AP margin. It shows that a multi-stage network is potential. However, the improvement becomes small when four or eight stages are employed. This indicates the necessity of a more effective single-stage module. Our single-stage model is discussed in Section~\ref{part:single_stage_module} and the performance with 71.5 AP on minival dataset demonstrates the superiority of our single-stage module. And our 2-stage network further leads to a 3.0 improvement and obtains 74.5 AP. Introducing the third and fourth stage maintains a tremendous upward trend and eventually brings an impressive performance boost of 1.4 AP improvement. With similar FLOPs, Hourglass has 71.3 AP at 4 stages and 71.6 at 8 stages, only 0.3 point. These experiments indicate that MSPN successfully pushes the upper bound of existing single-stage and multi-stage networks. It obtains noticeable performance gain with more network capacity.

\begin{table}[h]
	\begin{spacing}{1.1}
		\begin{center}
			\begin{tabular}{c|c c|c c}
				\hline
				Method & Res-50 & 2$\times$Res-18 & L-XCP & 4$\times$ S-XCP \\
				\hline
				AP & 71.5 & 71.6 & 73.7 & 74.7\\
				FLOPs & 4.4G & 4.0G & 6.1G & 5.7G\\
				\hline
			\end{tabular}
		\end{center}
		\caption{Results of MSPN with smaller single-stage modules on COCO minival dataset. "L-XCP" and "S-XCP" respectively represent a small and a large Xception backbone.}
		\label{tab:small}
	\end{spacing}
\end{table}

Finally, we testify that our approach is general. The down sampling unit of the single-stage module can effectively adopt other backbones. To verify that, we compare the proposed multi-stage network against any single-stage one with similar FLOPs. We conduct more experiments on ResNet-18 and Xception~\cite{chollet2017xception} architectures. Results are illustrated in Table~\ref{tab:small}. It is clear that the 2-stage network based on Res-18 obtains a comparable result with Res-50 with smaller FLOPs. Moreover, we design two Xception~\cite{chollet2017xception} backbones with different capacity, a large one~(L-XCP) and a small one~(S-XCP). The 4-stage S-XCP outperforms the single large model with 1.0 in AP with similar complexity. These results demonstrate the generality of our single-module backbone.

\subsubsection{Cross Stage Feature Aggregation}

To address the issue that a deep multi-stage architecture is vulnerable by information losing during repeated up and down sampling procedures, we propose a cross stage feature aggregation strategy. It is adopted to fuse different level features in adjacent stages and ensure more discriminative representations for the current stage. Table~\ref{tab:ablation} shows that the proposed feature aggregation strategy brings about a 0.3 gain from 74.2 to 74.5 for MSPN and a 0.5 improvement in terms of Hourglass, which demonstrates its effectiveness on dealing with aforementioned problems. At the same time, we can draw a conclusion that Hourglass tends to lose more information during forwarding propagation and our feature aggregation strategy can effectively mitigate this issue.

\begin{table}[h]
	\begin{center}
		\begin{threeparttable}
			\begin{tabular}{c c c | c | c}
				\hline
				\multicolumn{3}{c|}{\bf Components} & \multirow{2}*{\bf Hourglass} & \multirow{2}*{\bf MSPN}\\
				BaseNet & CTF & CSFA & ~ & ~ \\
				\hline
				$\surd$ & & & 71.3 & 73.3 \\
				$\surd$ & $\surd$ & & 72.5 & 74.2\\
				$\surd$ & $\surd$ & $\surd$ & 73.0 & 74.5\\
				\hline
			\end{tabular}
		\end{threeparttable}
	\end{center}
	\caption{Ablation Study of MSPN on COCO minival dataset. 'BaseNet' represents a 4-stage Hourglass or 2-stage MSPN based on Res-50 with similar complexity, see Table~\ref{tab:multi}. 'CTF' indicates the coarse-to-fine supervision. 'CSFA' means the cross stage feature aggregation.}
	\label{tab:ablation}
\end{table}

\subsubsection{Coarse-to-fine Supervision}

In this part, we evaluate our coarse-to-fine supervision for both MSPN and Hourglass. The results are shown in Table~\ref{tab:ablation}. It is clear that this strategy improves the performance of our network by a large margin from 73.3 to 74.2. First of all, this design aims to realize a coarse-to-fine detection procedure and the result demonstrates its effectiveness on further improving the accuracy of keypoint localization. In addition, it is reasonable that intermediate supervisions can take full advantage of contextual information across different scales. To demonstrate the applicability of this supervision in other multi-stage networks, we further apply this strategy to a 4-stage Hourglass that is comparable with our 2-stage MSPN in complexity, and finally obtains a 1.2 improvement in AP. In a word, the proposed coarse-to-fine supervision could largely boost the performance of pose estimation and be well adapted to other multi-stage networks.

Furthermore, we conduct several experiments to verify which level of supervision will have higher efficiency in our network. As described in Section~\ref{part:csfa}, we apply a Gaussian blur operation to each point on a heat map and a smaller kernel corresponds to a finer supervision. As shown in Table~\ref{tab:supervision}, we could see that either setting-1 or setting-2 will degrade the performance compared with the proposed coarse-to-fine supervision~(setting-3). Especially, setting-2 even leads to a worse performance than the setting-1, which indicates that an appropriate supervision could make a difference to the final result.

\begin{table}[h]
	\begin{spacing}{1.1}
		\begin{center}
			\begin{tabular}{c|c c c}
				\hline
				Setting & 1 & 2 & 3 \\
				\hline
				Kernel Size 1  & 7 & 5 & 7 \\
				Kernel Size 2  & 7 & 5 & 5 \\
				\hline
				AP & 74.2 & 74.0 & 74.5 \\
				\hline
			\end{tabular}
		\end{center}
		\caption{Results of a 2-stage MSPN with different supervision strategies on COCO minival dataset. The kernel size controls the fineness of supervision and a smaller value indicates a finer setting.}
		\label{tab:supervision}
	\end{spacing}
\end{table}

\begin{table}[h]
	\begin{spacing}{1.1}
		\begin{center}
			\begin{tabular}{c|c c c}
				\hline
				Detector & CPN(41.1) & Ours(49.4) & GT \\
				\hline
				2-Stg MSPN & 74.1 & 74.5 & 75.1 \\
				3-Stg MSPN & 74.8 & 75.2 & 75.6 \\
				4-Stg MSPN & 75.4 & 75.9 & 76.5 \\
				\hline
			\end{tabular}
		\end{center}
		\caption{Results of MSPN using three detectors on COCO minival dataset.}
		\label{tab:detection}
	\end{spacing}
\end{table}

\subsection{Influence of Human Detector}

\begin{table*}[t]
	\begin{spacing}{1.05}
		\begin{center}
			\renewcommand\tabcolsep{4.0pt}
			\begin{tabular}{l|c|c|c c c c c c c c c c}
				\hline
				Method & Backbone & Input Size & AP & AP$^{50}$ & AP$^{75}$ & AP$^{M}$ & AP$^{L}$ & AR & AR$^{50}$ & AR$^{75}$ & AR$^{M}$ & AR$^{L}$\\
				\hline
				CMU Pose~\cite{cao2016realtime} & - & - & 61.8 & 84.9 & 67.5 & 57.1 & 68.2 & 66.5 & 87.2 & 71.8 & 60.6 & 74.6\\
				Mask R-CNN~\cite{he2017mask} & Res-50-FPN & - & 63.1 & 87.3 & 68.7 & 57.8 & 71.4 & - & - & - & - & -\\
				G-RMI~\cite{papandreou2017towards} & Res-152 & 353$\times$257 & 64.9 & 85.5 & 71.3 & 62.3 & 70.0 & 69.7 & 88.7 & 75.5 & 64.4 & 77.1\\
				AE~\cite{newell2017associative} & - & 512$\times$512 & 65.5 & 86.8 & 72.3 & 60.6 & 72.6 & 70.2 & 89.5 & 76.0 & 64.6 & 78.1\\
				CPN~\cite{chen2018cascaded} & Res-Inception & 384$\times$288 & 72.1 & 91.4 & 80.0 & 68.7 & 77.2 & 78.5 & 95.1 & 85.3 & 74.2 & 84.3\\
				Simple Base~\cite{xiao2018simple} & Res-152 & 384$\times$288 & 73.7 & 91.9 & 81.1 & 70.3 & 80.0 & 79.0 & - & - & - & -\\
				HRNet ~\cite{sun2019deep} & HRNet-W48 & 384$\times$288 & 75.5 & 92.5 & 83.3 & 71.9 & 81.5 & 80.5 & - & - & - & - \\
				\textbf{Ours~(MSPN)} & 4$\times$Res-50 & 384$\times$288 & \textbf{76.1} & \textbf{93.4} & \textbf{83.8} & \textbf{72.3} & \textbf{81.5} & \textbf{81.6} & \textbf{96.3} & \textbf{88.1} & \textbf{77.5} & \textbf{87.1}\\
				\hline
				CPN+~\cite{chen2018cascaded} & Res-Inception & 384$\times$288 & 73.0 & 91.7 & 80.9 & 69.5 & 78.1 & 79.0 & 95.1 & 85.9 & 74.8 & 84.6\\
				Simple Base+*~\cite{xiao2018simple} & Res-152 & 384$\times$288 & 76.5 & 92.4 & 84.0 & 73.0 & 82.7 & 81.5 & 95.8 & 88.2 & 77.4 & 87.2\\
				HRNet* ~\cite{sun2019deep} & HRNet-W48 & 384$\times$288 & 77.0 & 92.7 & 84.5 & 73.4 & 83.1 & 82.0 & - & - & - & - \\
				Ours~(MSPN*) & 4$\times$Res-50 & 384$\times$288 & 77.1 & 93.8 & 84.6 & 73.4 & 82.3 & 82.3 & 96.5 & 88.9 & 78.4 & 87.7\\
				\textbf{Ours~(MSPN+*)} & 4$\times$Res-50 & 384$\times$288 & \textbf{78.1} & \textbf{94.1} & \textbf{85.9} & \textbf{74.5} & \textbf{83.3} & \textbf{83.1} & \textbf{96.7} & \textbf{89.8} & \textbf{79.3} & \textbf{88.2}\\ 
				\hline
			\end{tabular}
		\end{center}
		\caption{Comparisons of results on COCO test-dev dataset. "+" indicates using an ensemble model and "*" means using external data.}
		\label{tab:test-dev}
	\end{spacing}
\end{table*}

\begin{table*}[t]
	\begin{spacing}{1.05}
		\begin{center}
			\renewcommand\tabcolsep{4.0pt}
			\begin{tabular}{l|c|c|c c c c c c c c c c}
				\hline
				Method & Backbone & Input Size & AP & AP$^{50}$ & AP$^{75}$ & AP$^{M}$ & AP$^{L}$ & AR & AR$^{50}$ & AR$^{75}$ & AR$^{M}$ & AR$^{L}$\\
				\hline
				Mask R-CNN*~\cite{he2017mask} & ResX-101-FPN & - & 68.9 & 89.2 & 75.2 & 63.7 & 76.8 & 75.4 & 93.2 & 81.2 & 70.2 & 82.6\\
				G-RMI*~\cite{papandreou2017towards} & Res-152 & 353$\times$257 & 69.1 & 85.9 & 75.2 & 66.0 & 74.5 & 75.1 & 90.7 & 80.7 & 69.7 & 82.4\\
				CPN+~\cite{chen2018cascaded} & Res-Inception & 384$\times$288 & 72.1 & 90.5 & 78.9 & 67.9 & 78.1 & 78.7 & 94.7 & 84.8 & 74.3 & 84.7\\
				Sea Monsters+* & - & - & 74.1 & 90.6 & 80.4 & 68.5 & 82.1 & 79.5 & 94.4 & 85.1 & 74.1 & 86.8\\
				Simple Base+*~\cite{xiao2018simple} & Res-152 & 384$\times$288 & 74.5 & 90.9 & 80.8 & 69.5 & 82.9 & 80.5 & 95.1 & 86.3 & 75.3 & 87.5\\
				\textbf{Ours~(MSPN+*)} & 4$\times$Res-50 & 384$\times$288 & \textbf{76.4} & \textbf{92.9} & \textbf{82.6} & \textbf{71.4} & \textbf{83.2} & \textbf{82.2} & \textbf{96.0} & \textbf{87.7} & \textbf{77.5} & \textbf{88.6}\\
				\hline
			\end{tabular}
		\end{center}
		\caption{Comparisons of results on COCO test-challenge dataset. "+" means using an ensemble model and "*" means using external data.}
		\label{tab:test-challenge}
	\end{spacing}
\end{table*}

The human detector used in this work has a strong performance. It has 49.4 AP on COCO minival dataset. To evaluate its influence on the final pose estimation accuracy, we also test another detector with worse performance (the one in CPN~\cite{chen2018cascaded} with 41.1 AP) and an ``oracle detector'' using ground truth boxes, for controlled comparison. Pose estimation performance is reported in Table~\ref{tab:detection}. Clearly, with a much better detector, the pose estimation accuracy is only slightly improved. For example, there is only 0.5 gain from 41.1 detector to 49.4 one using 4-stage MSPN. This verifies that the good performance mostly comes from MSPN. The influence of detector is quite limited. The same conclusion is also drawn in ~\cite{chen2018cascaded}. Note that all results of Hourglass, ResNet and Xception in Section~\ref{part:ablation} are based on the same 49.4 detector.

\subsection{Comparison with State-of-the-art Methods}
\label{part:sota}

On COCO benchmark, as shown in Table~\ref{tab:test-dev}, our single model trained by only COCO data achieves 76.1 AP on test-dev and outperforms other methods by a large margin in all metrics. Advocated by external data, MSPN leads to a 1.0 improvement resulting in 77.1 AP. And the ensemble model finally obtains 78.1. From Table~\ref{tab:test-challenge}, it is clear that our approach obtains 76.4 AP on the test-challenge dataset and shows its significant superiority over other state-of-the-art methods. Eventually, our method surpasses COCO 2017 Challenge winner CPN~\cite{chen2018cascaded} and Sample Baseline~\cite{xiao2018simple} by 4.3 and 1.9 AP in test-challenge dataset respectively.

\begin{table}[t]
	\small
	\begin{spacing}{1.2}
		\begin{center}
			\renewcommand\tabcolsep{2.5pt}
			\begin{tabular}{l|c c c c c c c | c}
				\hline
				Method & Hea & Sho & Elb & Wri & Hip & Kne & Ank & Mean\\
				\hline
				Bulat \etal~\cite{bulat2016human} & 97.9 & 95.1 & 89.9 & 85.3 & 89.4 & 85.7 & 81.7 & 89.7 \\
				Newell \etal~\cite{newell2016stacked} & 98.2 & 96.3 & 91.2 & 87.1 & 90.1 & 87.4 & 83.6 & 90.9 \\
				Tang \etal~\cite{tang2018quantized} & 97.4 & 96.4 & 92.1 & 87.7 & 90.2 & 87.7 & 84.3 & 91.2 \\
				Ning \etal~\cite{ning2018knowledge} & 98.1 & 96.3 & 92.2 & 87.8 & 90.6 & 87.6 & 82.7 & 91.2 \\
				Luvizon \etal~\cite{luvizon2017human} & 98.1 & 96.6 & 92.0 & 87.5 & 90.6 & 88.0 & 82.7 & 91.2 \\
				Chu \etal~\cite{Xiao-Multi} & 98.5 & 96.3 & 91.9 & 88.1 & 90.6 & 88.0 & 85.0 & 91.5 \\
				Chou \etal~\cite{chou2017self} & 98.2 & 96.8 & 92.2 & 88.0 & 91.3 & 89.1 & 84.9 & 91.8 \\
				Chen \etal~\cite{chen2017adversarial} & 98.1 & 96.5 & 92.5 & 88.5 & 90.2 & 89.6 & 86.0 & 91.9 \\
				Yang \etal~\cite{yang2017learning} & 98.5 & 96.7 & 92.5 & 88.7 & 91.1 & 88.6 & 86.0 & 92.0 \\
				Ke \etal~\cite{ke2018multi} & 98.5 & 96.8 & 92.7 & 88.4 & 90.6 & 89.3 & 86.3 & 92.1 \\
				Tang \etal~\cite{tang2018deeply} & 98.4 & 96.9 & 92.6 & 88.7 & 91.8 & 89.4 & 86.2 & 92.3 \\
				Sun \etal~\cite{sun2019deep} & \textbf{98.6} & 96.9 & 92.8 & 89.0 & 91.5 & 89.0 & 85.7 & 92.3 \\
				Zhang \etal~\cite{zhang2019human} & \textbf{98.6} & 97.0 & 92.8 & 88.8 & 91.7 & 89.8 & \textbf{86.6} & 92.5 \\
				\hline
				\textbf{Ours~(MSPN)} & 98.4 & \textbf{97.1} & \textbf{93.2} & \textbf{89.2} & \textbf{92.0} & \textbf{90.1} & 85.5 & \textbf{92.6} \\
				\hline
			\end{tabular}
		\end{center}
		\caption{Comparisons of results on MPII test dataset.}
		\label{tab:mpii}
	\end{spacing}
\end{table}

MPII is another popular benchmark for pose estimation. We also validate the proposed MSPN on this dataset. The PCKh@0.5 result on MPII test dataset is shown in Table~\ref{tab:mpii}. Our result is the new state-of-the-art\footnote{http://human-pose.mpi-inf.mpg.de/$\sharp$results}.

\begin{figure*}[t]
	\begin{center} 
		\includegraphics[width=0.80\linewidth]{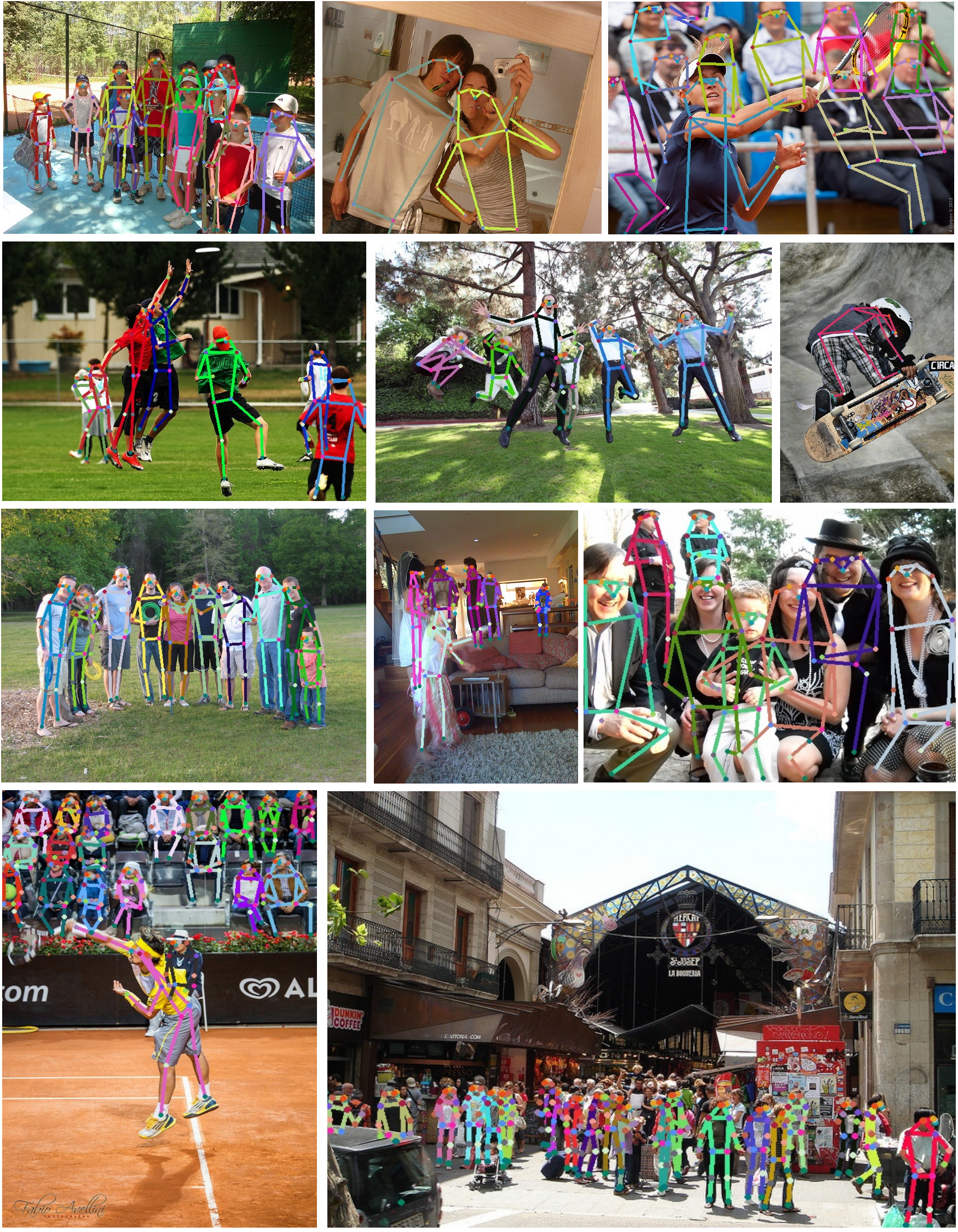} 
	\end{center} 
	\caption{Visualization of MSPN results on COCO minival dataset.} 
	\label{fig:examples} 
\end{figure*} 

Some results generated by our method are shown in Figure~\ref{fig:examples}. We can see that our MSPN handles crowd and occlusion situations as well as challenging poses effectively.

\section{Conclusion}
In this work, we propose a Multi-Stage Pose Network~(MSPN) to perform multi-person pose estimation. We first verify the effectiveness of the multi-stage pipeline with well-designed single-stage modules in MSPN. Additionally, a coarse-to-fine supervision and a cross stage feature aggregation strategy are proposed to further boost the performance of our framework. Extensive experiments have been conducted to demonstrate its effectiveness. For the first time, it is shown that a multi-stage architecture is competitive on the challenging COCO dataset.

{\small
	\bibliographystyle{ieee}
	\bibliography{egbib}
}

\end{document}